\documentclass[letterpaper, 10 pt, conference]{ieeeconf} 

\IEEEoverridecommandlockouts    

\let\labelindent\relax

\usepackage{graphicx}
\usepackage{graphics}
\usepackage{pdfpages}
\usepackage{epsfig}
\usepackage{amsmath}
\usepackage{amssymb} 
\usepackage{amsfonts}
\usepackage{cite}
\usepackage{stfloats}
\graphicspath{{Fig/}}
\usepackage{multirow}
\usepackage{algorithm} 
\usepackage{algpseudocode} 
\usepackage{enumitem}

\usepackage{amsthm} 
\usepackage{geometry}
\usepackage[capitalize]{cleveref}

\title{\LARGE \bf
Design and Control of a Perching Drone Inspired by the Prey-Capturing Mechanism of Venus Flytrap 
}

\author{Ye Li$^{1}$,\textit{ Graduate Student Member, IEEE}, Daming Liu$^{1}$, Yanhe Zhu$^*$$^{1}$,\textit{ Member, IEEE}, Junming Zhang$^*$$^{1}$, \\Yongsheng Luo$^{1}$, Ziqi Wang$^{1}$, Chenyu Liu$^{1}$, Jie Zhao$^{1}$,\textit{ Senior Member, IEEE} 
\thanks{This Work was supported by the National Science Fund for Distinguished Young Scholars of China under Grant 52025054 and the Key Program of the National Natural Science Foundation of China under Grant 52435001.\textit{(Ye Li and Daming Liu contributed equally to this work. Corresponding author: Yanhe Zhu, Junming Zhang.)}}
\thanks{$^{1}$The authors are with the State Key Laboratory of Robotics and Systems, Harbin Institute of Technology, Harbin 150000, China.}
}

\begin{document}

\maketitle
\thispagestyle{empty}
\pagestyle{empty}

\begin{abstract}

The endurance and energy efficiency of drones remain critical challenges in their design and operation. To extend mission duration, numerous studies explored perching mechanisms that enable drones to conserve energy by temporarily suspending flight. This paper presents a new perching drone that utilizes an active flexible perching mechanism inspired by the rapid predation mechanism of the Venus flytrap, achieving perching in less than 100 ms. The proposed system is designed for high-speed adaptability to the perching targets. The overall drone design is outlined, followed by the development and validation of the biomimetic perching structure. To enhance the system stability,  a cascade extended high-gain observer (EHGO) based control method is developed, which can estimate and compensate for the external disturbance in real time. The experimental results demonstrate the adaptability of the perching structure and the superiority of the cascaded EHGO in resisting wind and perching disturbances.

\end{abstract}

\section{Introduction}

Drones, with their flexibility and wide working range, offer significant advantages in environmental exploration, high-altitude operations, and reconnaissance and rescue\cite{zheng2024new,dautzenberg2023perching,xu2024biomimetic}. To improve drone endurance, many researchers are inspired by birds to incorporate perching mechanisms. After perching, the drone can deactivate or partially deactivate the propeller, maintaining altitude, reducing rotor noise and extending endurance greatly\cite{hang2019perching,lan2024aerial,wuest2024agile}. However, the process of perching introduces physical interactions with the environment, and complex habitats and external disturbances further complicate the drone’s dynamic response and stability during perching. Therefore, developing an adaptive, responsive perching mechanism capable of accommodating targets of varying sizes is essential for improving mission success rates.

Drones can achieve perching either by utilizing their own arms or by integrating specialized perching devices. The use of drone’s arms eliminates the need for additional hardware and reduces weight, which requires the drone to have high flexibility and precise control\cite{tao2023design}. On the other hand, perching devices are generally classified into adhesion and grasping mechanisms. Adhesion  methods rely on negative pressure, electromagnetic forces, or static electricity to facilitate attachment \cite{li2022aerial,habas2025ceilings,luo2025high,graule2016perching,park2020lightweight}. Technologies such as vacuum suction cups and electrostatic adsorption allow drones to adhere to walls or ceilings. However, these methods require relatively flat and smooth surfaces. Recent developments, including hybrid suction cups and high-frequency voltage-driven adhesion, improve adaptability to certain extent. However, adhesion stability remains a challenge, particularly when surface roughness varies. In contrast, grasping mechanisms are better suited for unstructured environments and can be divided into active and passive approaches. Active grasping mechanisms utilize controllable structures to achieve perching, but their response speed is inherently limited by mechanical constraints \cite{kitchen2020design,iida2025adaptive}. Passive grasping mechanisms rely on collision energy to trigger rapid closure, enabling quick response times. However, these mechanisms require precise control over the drone’s collision posture, which can impact stability during perching \cite{firouzeh2023perching,roderick2021bird}. Compared to rigid perching mechanisms, flexible perching mechanisms offer greater adaptability, providing a larger contact area and a more secure wrapping effect \cite{ching2024crawling,guo2024powerful,chen2024scale}.

\begin{figure}[t!] 
	\centerline{\includegraphics[width=8.5cm]{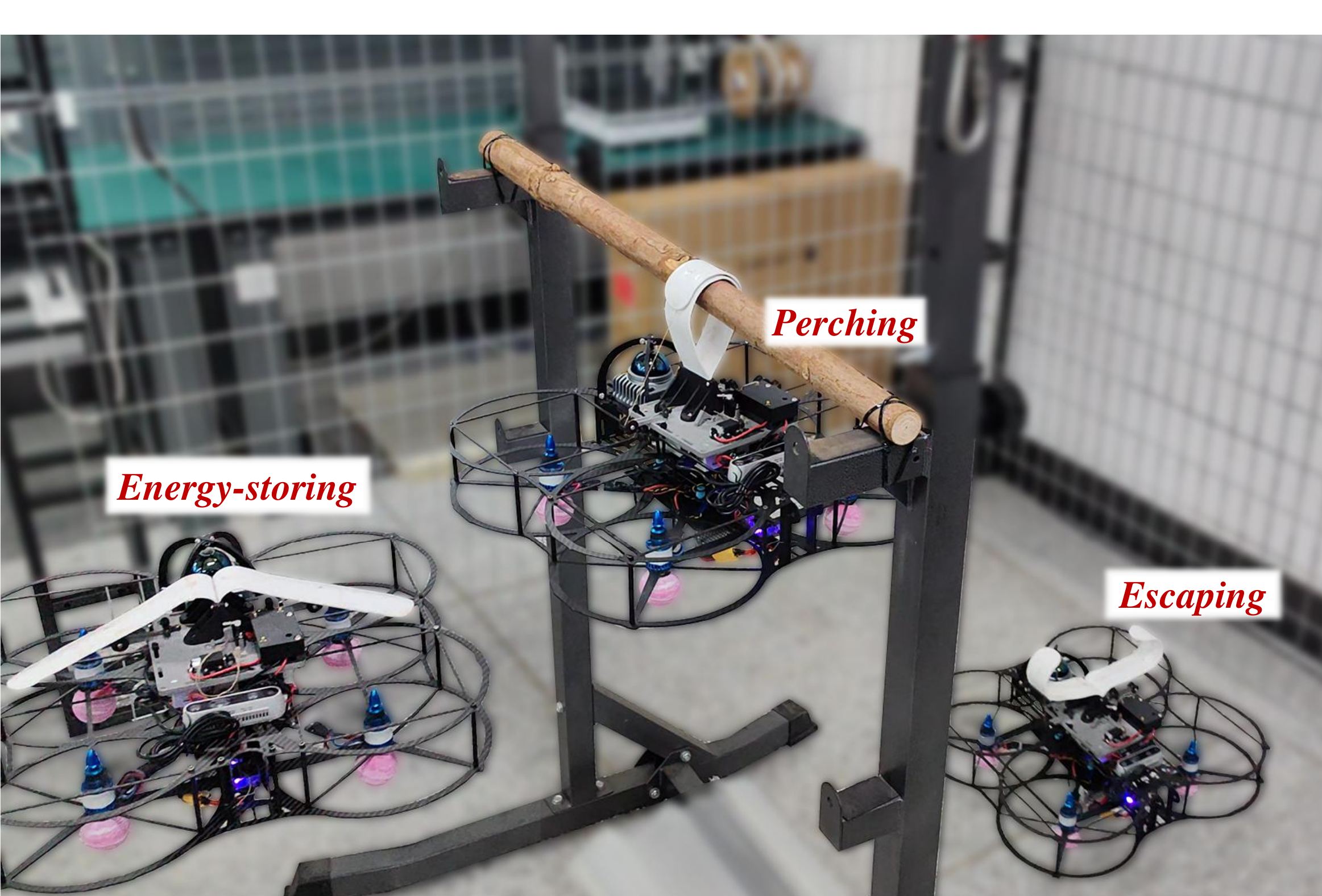}}
 \caption{Three states of the drone in the perching mission.}
	\label{fig:overview}
\end{figure}\normalsize

In summary, perching mechanisms offer greater adaptability across a diverse range of perching targets. Active perching mechanisms provide direct control over the perching process. However, their effectiveness is constrained by mechanical response time. Passive perching mechanisms, in contrast, enable rapid closure through collision-triggered activation, but they require precise impact control, imposing stringent accuracy demands on the drone’s navigation and stabilization. Given these considerations, the development of a flexible, active, rapid-response, and stable perching mechanism is essential for enhancing drone perching capabilities and reducing energy consumption, particularly in complex and unstructured environments.

This paper makes the following key contributions: 1) A flexible perching mechanism inspired by the Venus flytrap is designed, integrating both proactiveness and high-speed response. The mechanism achieves perching within 100 ms and adapts to varying target sizes and shapes within a defined range; 2) A new cascaded extended high-gain observer (EHGO) is developed to estimate and compensate for disturbances encountered during both flight and perching, significantly improving drone stability. The remainder of this paper is structured as follows: Section \uppercase\expandafter{\romannumeral2} details the mechanical and hardware design of the perching drone, Section \uppercase\expandafter{\romannumeral3} presents the drone modeling and control framework, Section \uppercase\expandafter{\romannumeral4} discusses the design of the cascaded EHGO along with its stability proof, Section \uppercase\expandafter{\romannumeral5} provides experimental validation and performance analysis, and Section \uppercase\expandafter{\romannumeral6} summarizes key results and outlining potential future research directions.

\begin{figure*}[t!]
\centering
 {\includegraphics[width=17.7cm]{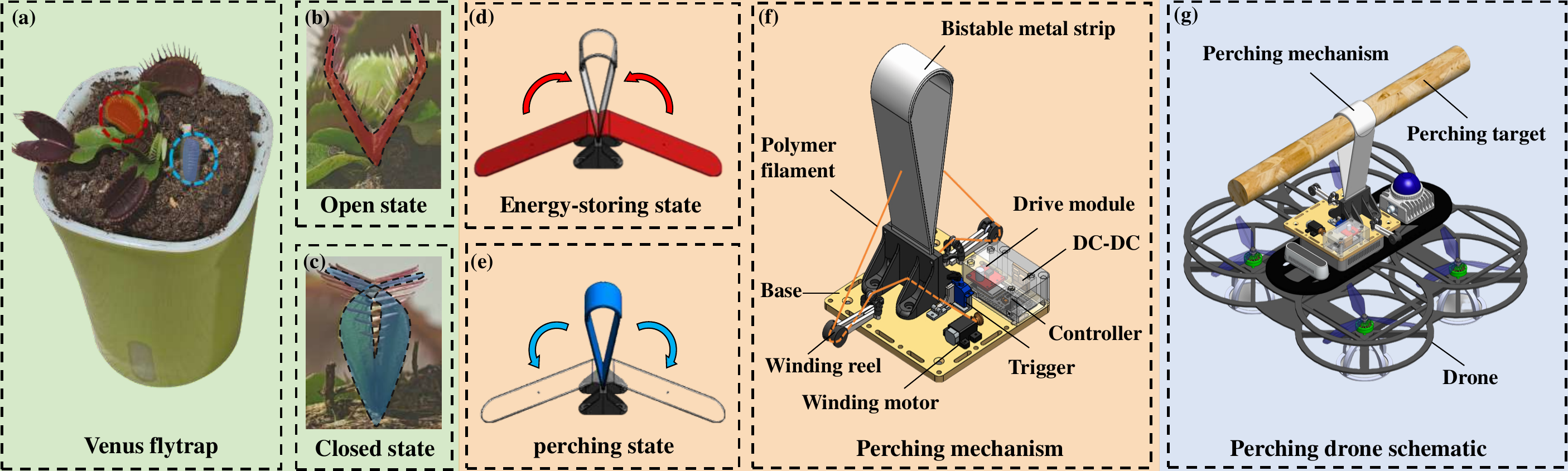}}
\caption{Concept and implementation for bioinspired perching mechanism: (a) Venus flytrap in its natural state; (b) Open state of the Venus flytrap before prey capture; (c) Closed state of the Venus flytrap after triggering; (d) Energy-storing phase of the perching mechanism, primed for rapid closure; (e) Perching state of the mechanism, enabling the drone to release by opening the perching strips; (f) Venus flytrap-inspired perching mechanism, designed for rapid actuation upon signal activation; (g) Drone equipped with the perching mechanism, demonstrating its capability to grasp cylindrical targets.}
\label{机构机理图}
\end{figure*}

\section{Mechanical Design}
This section mainly introduces the design of the perching mechanism of the drone inspired by the Venus flytrap and the hardware integration of the overall system.
\label{sec:DESIGN}

\subsection{Perching Mechanism Biomimetic Characteristics}

The Venus flytrap is capable of closing its leaves within 100 ms, earning it the nickname “plant hunter”. The rapid closure of the Venus flytrap in its natural state is driven by the elastic potential energy storage and release mechanism inherent in its bistable leaf structure.This biological process is characterized by three key features: preloaded energy storage, trigger-activated rapid release, and enhanced surface adhesion. Inspired by the Venus flytrap, this study presents an ultrafast active perching mechanism. The specific design is shown in \cref{机构机理图}. In terms of energy storage, the proposed mechanism employs a bistable metal band to emulate the mechanical behavior of the flytrap’s leaves. When the mechanism is in the energy-storing state, elastic potential energy is preloaded via a tensioning system—a process closely analogous to the biological energy accumulation in the Venus flytrap through cell turgor pressure. The triggering mechanism is designed based on the principle of action potential conduction in the flytrap. By using a servo-driven clamp to release a pre-compressed spring, the system achieves a rapid closure response within 100 ms, a performance metric that parallels the natural prey-capturing speed of the plant. Furthermore, the surface functionalization of the mechanism is inspired by the mucilage secretion mechanism of the Venus flytrap. Coating the metal bands with silicone not only improves grasping stability but also enables the mechanism to accommodate a wide range of objects. This multi-scale biomimetic approach ensures that the mechanism retains its high-speed responsiveness while achieving adaptability and reliability comparable to its biological counterpart.
\subsection{Perching Mechanism Design}
The perching mechanism is composed of perching strips, pull-wire mechanism, locking-trigger mechanism, and the electronic control unit. The tensioning mechanism is powered by a GA12-N20 reduction motor (12 V, 30 rpm). The tensioning rope follows a structured pathway: first passing through the locking-release mechanism, then extending from both sides of the perching belt base, and finally connecting to the perching belt using guide wheels. By controlling the rotation of the reduction motor, the rope is wound onto the reel, mechanically applying force to open the perching belt and transition it into a tensioned state, thereby preparing the drone for escape after perching. The locking-release device consists of a mechanical assembly and a servo control unit. The servo used is a PZ-15318 ultra-miniature digital servo (2.2 g), which operates in conjunction with a caliper and a spring. Once the tensioning mechanism pulls apart the bistable metal strips on both sides, the caliper clamps the passing tensioning rope, maintaining the perching mechanism in a tensioned state through servo-driven actuation. When the drone initiates perching, the servo rotates, releasing the compressed spring within the caliper. As the spring returns to its original position, the bistable structure releases its stored elastic energy, triggering a rapid grasping action to achieve perching.

\begin{figure}[t] 
	\centerline{\includegraphics[width=8.6cm]{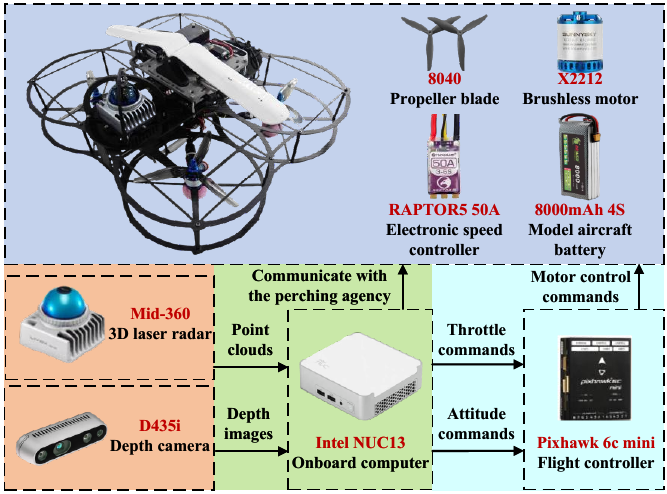}}
 \caption{Hardware integration and information flow in the perching drone.}
	\label{无人机系统图}
\end{figure}\normalsize

\subsection{Hardware Integration}
The hardware integration of the perching drone is illustrated in \cref{无人机系统图}. The total weight of the system is approximately 2.7 kg, which includes a LiPo battery (4S, 8000 mAh). The main body is constructed from a carbon fiber frame with an axis distance of 333 mm. The drone is powered by four X2212 brushless DC motors (980 kV) and two pairs of 8040 propellers. For getting the real-time location, a Mid-360 laser radar is installed at the front, offering a wide field of view, while a D435i camera mounted on top obtains the relative pose of the target. The Intel NUC13 serves as the computational unit, executing key algorithmic modules, including IMU and radar fusion, visual recognition, dynamic analysis, and position control. The Pixhawk 6C, running PX4 firmware, handling attitude control and thrust management of the motors. The perching mechanism is mounted directly above the drone, ensuring alignment between the center of gravity of the drone and the mechanism. This configuration simplifies modeling and control while minimizing interference from rotor downwash airflow during perching.

\section{Drone Model and Control Problem}

Drones encounter multiple disturbances during flight and perching, including model inaccuracies, impact disturbances, wind disturbances, and rotor resistance. In flight, wind from various directions can alter the drone’s attitude and position, making accurate state estimation challenging. During perching, the perching mechanism must physically engage with the perching target, generating instantaneous disturbances that are difficult to predict. These disturbances can destabilize the drone’s attitude, potentially leading to instability or even failure to perch successfully. Let \(d_f\) and \(d_\tau\) denote the force and torque induced by disturbances, respectively. We assume that exist two constants \(\bar{d}_f\) and \(\bar{d}_\tau\), such that $\| \dot{d}_f \| \leq \bar{d}_f$ and $\| \dot{d}_\tau \| \leq \bar{d}_\tau$.

The body coordinate system and world coordinate system of the drone are defined in accordance with conventional aerospace and robotics frameworks. The specific coordinate system representation is depicted in \cref{实验轨迹}. By integrating the classic kinematic and dynamic models of the drone while incorporating the previously discussed disturbances, the governing equations for position kinematics and attitude dynamics can be formulated as follows:
\begin{equation}
\begin{cases}
m\dot{v} = F - mg e_3 + d_f\\
M(\eta)\ddot{\eta} + C(\eta,\dot{\eta})\dot{\eta} = \tau + d_{\tau}
\end{cases}
\end{equation}
where $m$ is the total mass of the drone; $F$ is the total lift provided by the propellers, $g$ is the gravitational acceleration, $M(\eta )\in {{\mathbb{R}}^{3\times 3}}$ is the inertia matrix, $C(\eta,\dot{\eta })\in {{\mathbb{R}}^{3\times 3}}$ is the centrifugal and Coriolis matrix, $\tau$ is the equivalent control moment of the four thrust forces on the drone; ${v}$ is the linear velocity of the drone; and $\eta ={{\left[ \phi,\theta,\psi \right]}^{\text{T}}}$ is the Euler angle between the world frame and the body frame.

The errors in the drone’s position, velocity, and angular orientation are defined as follows:
\begin{equation}
{{e}_{s}}={{s}_{d}}-s,\,{{e}_{v}}={{v}_{d}}-v, \,{{e}_{\eta }}={{\eta }_{d}}-\eta 
\end{equation}
where ${s}$ is the current trajectory and ${s}_{d}$, ${v}_{d}$, ${\eta}_{d}$ are the desired trajectory, velocity, and Euler angle of the drone, respectively.


To facilitate the design process of the cascaded EHGO, the position dynamics of the drone can be formulated as a second-order uncertain nonlinear system, expressed as follows:
\begin{equation}
\begin{cases}
{{{\dot{x}}}_{p1}}={{x}_{p2}} \\ 
{{{\dot{x}}}_{p2}}=\left(-g{{e}_{3}}+\frac{{{d}_{f}}}{m}\right)+\frac{1}{m}{{u}_{p}} \\ 
{y_{p}}={{x}_{p1}} \label{secobd-orderp}
\end{cases}
\end{equation}
where ${{x}_{p1}}={e}_{s}$, ${{x}_{p2}}={e}_{v}$ and ${{u}_{p}}$ is the expected control lift in the position loop.

The expression of attitude dynamics is similar to the position dynamics, which can be expressed as follows:
\begin{equation}
\begin{cases}
{{{\dot{x}}}_{a1}}={{x}_{a2}} \\ 
{{{\dot{x}}}_{a2}}={{M}^{-1}}\left( {{d}_{\tau}}-C{{x}_{a2}} \right)+{{M}^{-1}} \ {{u}_{a}} \\ 
{y_{a}}={{x}_{a1}} \label{secobd-ordera}
\end{cases}
\end{equation}
where ${{x}_{a1}}={{e}_{\eta }}$, ${{x}_{a2}}=\dot{e}_{\eta }$ and ${{u}_{a}}$ is the expected control torque in the attitude loop.


Building on the fundamental principles of EHGO-based control \cite{khalil2017extended,guo2013convergence,sariyildiz2019disturbance}, the extended states of the position loop and attitude loop, denoted by ${{x}_{p3}}$ and ${{x}_{a3}}$, are given by
\begin{equation}
{{x}_{p3}}=\left(-g{{e}_{3}}+\frac{{{d}_{f}}}{m}\right)+\left(\frac{1}{m}-\frac{1}{{{m}_{0}}}\right){{u}_{p}}\label{xp3}
\end{equation}
\begin{equation}
{{x}_{a3}}={{M}^{-1}}\left( {{d}_{\tau }}-C{{x}_{a2}} \right) 
+\left( {{M}^{-1}}-M_{0}^{-1} \right){{u}_{a}} \\ \label{xa3}
\end{equation}
where ${m}_{0}$ and ${M}_{0}$ are the known nominal mass and inertia matrix of the drone, respectively. By \eqref{xp3} and \eqref{xa3}, the controls are given by:
\begin{equation}
{{u}_{p}}={{B}_{p}}\cdot \text{sat}\left(\frac{{{m}_{0}}\left( K{{{\hat{x}}}_{p}}-{{{\hat{x}}}_{p3}} \right)}{{B}_{p}}\label{controlor}\right)
\end{equation}
\begin{equation}
{{u}_{a}}={{B}_{a}}\cdot \text{sat}\left(\frac{{{M}_{0}}\left( K{{{\hat{x}}}_{a}}-{{{\hat{x}}}_{a3}}\right)}{{B}_{a}}\right)
\end{equation}
where $K=\left[ {{k}_{1}},{{k}_{2}} \right]\in {{\mathbb{R}}^{1\times 2}}$ 
is selected such that the matrix $E=\left[ \begin{matrix}
 0 & 1 \\
 {{k}_{1}} & {{k}_{2}} \\
\end{matrix} \right]$ is Hurwitz, $B$ is the saturation
bound to prevent peaking in the system during the initial phase of the observer; and $\text{sat}(\cdot )$ is an odd, smooth, nondecreasing function defined as $\text{sat}(v)=v$ if $\left| v \right|\le 1$, $\text{sat}'\triangleq \frac{\text{dsat}(v)}{\text{d}v}\le 1$, and ${{\lim }_{v\to \infty }}\text{sat}(v)=v(1+h)$ with $0<h\le 1$.

By \eqref{xp3} and \eqref{xa3}, the traditional EHGO requires that the discrepancy between the priori estimation and the true value of the control system model be minimized to ensure that the control gain remains within acceptable limits. Maintaining this minimal discrepancy is critical for achieving accurate disturbance estimation and ensuring robust system performance. However, if the system model exhibits significant uncertainty, the resulting discrepancy may become unmanageable, leading to potential design failure of EHGO. This could compromise the stability of the control system.
 

\section{Control based on cascaded EHGO}

To mitigate disturbances arising from perching shock, model inaccuracies, and external environmental factors, this study introduces a new cascaded EHGO. The cascaded EHGO of the position loop is expressed as follows:
\begin{equation}
\begin{cases}
{{{\dot{\xi }}}_{p1}}=\frac{{{l}_{p1}}}{\varepsilon }{{e}_{p1}} \\ 
{{{\dot{\xi }}}_{p2}}=\frac{{{l}_{p2}}}{\varepsilon }{{e}_{p2}}+\frac{{{u}_{p}}}{{{m}_{0}}} \\
\end{cases}\label{new EHGOp}
\end{equation}
The cascaded EHGO of the attitude loop can be determined as follows:
\begin{equation}
\begin{cases}
{{{\dot{\xi }}}_{a1}}=\frac{{{l}_{a1}}}{\varepsilon }{{e}_{a1}} \\ 
{{{\dot{\xi }}}_{a2}}=\frac{{{l}_{a2}}}{\varepsilon }{{e}_{a2}}+M_{0}^{-1}{{u}_{a}} \\
\end{cases}\label{new EHGOa}
\end{equation}
where ${{l}_{p1}},{{l}_{p2},{{l}_{a1}},{{l}_{a2}}}>0$, $\hat{{x}}_{p}={{\left[ {{x}_{p1}},{{{\hat{x}}}_{p2}} \right]}^{\text{T}}}$, $\hat{{x}}_{a}={{\left[ {{x}_{a1}},{{{\hat{x}}}_{a2}} \right]}^{\text{T}}}$and
\begin{align*}
 &{{e}_{p1}}={y_{p}}-{{\xi }_{p1}}, \,{{e}_{p2}}={{\hat{x}}_{p2}}-{{\xi }_{p2}}\\
 &{{e}_{a1}}={y_{a}}-{{\xi }_{a1}}, \,{{e}_{a2}}={{\hat{x}}_{a2}}-{{\xi }_{a2}}
\end{align*}
The observer’s output is described as follows:
\begin{equation}
\begin{aligned}
&{{\hat{x}}_{p2}}=\frac{{{l}_{p1}}}{\varepsilon }{{e}_{p1}},\,{{\hat{x}}_{p3}}=\frac{{{l}_{p2}}}{\varepsilon }{{e}_{p2}}\\
&{{\hat{x}}_{a2}}=\frac{{{l}_{a1}}}{\varepsilon }{{e}_{a1}},\,{{\hat{x}}_{a3}}=\frac{{{l}_{a2}}}{\varepsilon }{{e}_{a2}}
\end{aligned}
\end{equation}

The new EHGO can be interpreted as the cascade connection of two first-order high-gain observers. Under the control framework based on the cascaded EHGO, the stability proofs of the position closed-loop system and the attitude closed-loop system follow similar methods. This paper takes the position loop system as an example to prove the convergence of the proposed observer.

\textit{Theorem}: Considering the closed-loop system formed through \eqref{secobd-orderp}, \eqref{controlor}, and \eqref{new EHGOp}, the saturation limit ${B}_{p}$ is appropriately selected. For any bounded initial state ${{x}_{p}}(0)$ and $T>0$, exist ${{\varepsilon }^{*}}>0$ such that for any $\varepsilon \in (0,{{\varepsilon }^{*}})$ and $t\in \left[ T,\infty \right)$, 
\begin{equation}
\left| {{x}_{pi}}(t)-{{{\hat{x}}}_{pi}}(t) \right|=O(\varepsilon ), i=2,3,
\end{equation}
\begin{equation}
{{\lim }_{t\to \infty }}\left\| {x}_{p}(t) \right\|=O\left( \varepsilon \right) \label{x shoulian}
\end{equation}

To prove the theorem, the boundedness of the system state ${x}_{p}$ is initially defined. Let $V({{x}_{p}})={{{{x}_{p}}}^{\text{T}}}P{{x}_{p}}$, where $P$ is the positive definite matrix solution to $PE+{{E}^{\text{T}}}P=-I$ with identity matrix $I$. For ${{\lambda }_{0}}=V({{x}_{p}}(0))+1$, two compact sets can be determined as follows:
\begin{equation}
{{\Omega }_{0}}=\left\{ {{x}_{p}}\in {{\mathbb{R}}^{2}}:V({{x}_{p}})\le {{\lambda }_{0}} \right\}
\end{equation}
\begin{equation}
{{\Omega }_{1}}=\left\{ {{x}_{p}}\in {{\mathbb{R}}^{2}}:V({{x}_{p}})\le {{\lambda }_{0}}+1 \right\}
\end{equation}
where ${{\Omega }_{0}}$ is a subset of ${{\Omega }_{1}}$ and ${{x}_{p}}(0)$ is in the ${{\Omega }_{0}}$. The next lemma shows that for a sufficiently high gain $1/\varepsilon $, the state of the system ${{x}_{p}}(t)$ remains in the set ${{\Omega }_{1}}$, for any $t\in \left[ 0,\infty \right)$.

\textit{Lemma}: Consider the closed-loop system formed through \eqref{secobd-orderp}, \eqref{controlor}, and \eqref{new EHGOp}, and the saturation limit ${B}_{p}$ is appropriately selected. For any bounded initial state ${{x}_{p}}(0)$, there exists ${{\varepsilon }^{*}}>0$ such that for any $\varepsilon \in \left( 0,{{\varepsilon }^{*}} \right)$ and $t\in \left[ 0,\infty \right)$, ${{x}_{p}}(t)\in {{\Omega }_{1}}$. 

\textit{Proof of the lemma}: Since the boundedness of ${{u}_{p}}$ and the fact that ${{x}_{p}}(0)$ is an interior point of ${{\Omega }_{0}}$, it can be easily verified that there exists ${{t}_{0}}>0$, which is independent of $\varepsilon $ such that ${{x}_{p}}(t)\in {{\Omega }_{0}}$, $\forall t\in \left[ 0,{{t}_{0}} \right]$.

The lemma will be proved by contradiction. Assume the lemma is false. Then there exists ${{t}_{2}}>{{t}_{1}}>{{t}_{0}}$ such that
\begin{equation}\label{lamuda assum}
\begin{cases}
V({{x}_{p}}({{t}_{1}}))={{\lambda }_{0}} \\
V({{x}_{p}}({{t}_{2}}))={{\lambda }_{0}}+1 \\
 {{\lambda }_{0}}\le V({{x}_{p}}(t))\le {{\lambda }_{0}}+1, t\in \left[ {{t}_{1}},{{t}_{2}} \right] \\
 V({{x}_{p}}(t))\le {{\lambda }_{0}}+1,t\in \left[ 0,{{t}_{2}} \right] \\ 
\end{cases}
\end{equation}

The cascaded EHGO estimation error ${\mu }_{p} ={{\left[ {{\mu }_{p1}},{{\mu }_{p2}} \right]}^{\text{T}}}$ can be reformulated as follows: 
\begin{align*}
{{\mu }_{p1}}={{x }_{p2}}-{{\hat{x }}_{p2}},\,\,{{\mu }_{p2}}=\varepsilon \left( {{x}_{p3}}-{{{\hat{x}}}_{p3}} \right)
\end{align*}
Subsequently, the convergence of ${{\mu }_{p1}}$, ${{\mu }_{p2}}$ and $x$ within the time interval $\left[ 0,{{t}_{2}} \right]$ is analyzed.

\textit{Convergence of ${{\mu }_{p1}}$}: Based on \eqref{secobd-orderp}, and \eqref{new EHGOp}, the dynamics of ${{\mu }_{p1}}$ can be expressed as follows: 

\begin{equation}
{{\dot{\mu }}_{p1}}=-\frac{{{l}_{1}}}{\varepsilon }{{\mu }_{p1}}+{{\dot{x}}_{p2}} \label{miu1 dao}
\end{equation}

Since ${{\dot{x}}_{p2}}=\left(-g{{e}_{3}}+\frac{{{d}_{f}}}{m}\right)+\frac{1}{m}{{u}_{p}}$ is bounded within the time interval $\left[ 0,{{t}_{2}} \right]$, there exists ${{\varepsilon }_{1}}>0$ such that any $\varepsilon \in (0,{{\varepsilon }_{1}})$ and ${{T}_{1}}\in ({{t}_{0}},{{t}_{1}})$

\begin{equation}
{{\mu }_{p1}}=O(\varepsilon ),\,\,\forall t\in \left[ {{T}_{1}},{{t}_{2}} \right]\label{mu1}
\end{equation}

\textit{Convergence of ${{\mu }_{p2}}$}: The dynamics of ${{\mu }_{p2}}$ can be determined as follows:
\begin{equation}
\begin{aligned}
\dot{\mu}_{p2} &= \varepsilon (\dot{x}_{p3} - \dot{\hat{x}}_{p3}) \\
&= \varepsilon \left( \frac{\dot{d}_f}{m} + \left( \frac{1}{m} - \frac{1}{m_0} \right) \dot{u} - \dot{\hat{x}}_{p3} \right)\label{miu2 dao}
\end{aligned}
\end{equation}

Since the mass of the drone $m>0$, and ${{d}_{f}}$, ${{\dot{d}}_{f}}$ and $(\frac{1}{m}-\frac{1}{{{m}_{0}}})$ are bounded by positive numbers independent of $\varepsilon $ within the time interval $\left[ 0,{{t}_{2}} \right]$, where the derivative of ${{u}_{p}}$ is given by

\begin{equation}
{{\dot{u}}_{p}}=\text{sa{t}}'\cdot {{{m}}_{0}}({{k}_{1}}{{x}_{p2}}+{{k}_{2}}{{\dot{\hat{x}}}_{p2}}-{{\dot{\hat{x}}}_{p3}}) \label{u dao}
\end{equation}

Through the convergence of ${{\mu }_{p1}}$ and the fact that ${{\dot{\hat{x}}}_{p2}}$ is restricted by a positive number independent of $\varepsilon $ within the time interval $\left[ {{T}_{1}},{{t}_{2}} \right]$. In \eqref{miu1 dao}, the dynamics of ${{\dot{\hat{x}}}_{p3}}$ can be expressed as follows:
\begin{equation}
\begin{aligned}
 {{{\dot{\hat{x}}}}_{p3}}&=\frac{{{l}_{p2}}}{\varepsilon }({{{\dot{\hat{x}}}}_{p2}}-{{{\dot{\mu }}}_{p2}}) \\ 
 & =\frac{{{l}_{p2}}}{{{\varepsilon }^{2}}}{{\mu }_{p2}}+\frac{{{l}_{p1}}{{l}_{p2}}}{{{\varepsilon }^{2}}}{{\mu }_{p1}}-\frac{{{l}_{p2}}}{{{\varepsilon }^{{}}}}{{{\dot{x}}}_{p2}} \label{x3 guji dao}
\end{aligned}
\end{equation}
Based on \eqref{miu2 dao}--\eqref{x3 guji dao}, the dynamics of ${{\mu }}_{p2}$ can be determined as follows:
\begin{equation}
\begin{aligned}
{{\dot{\mu }}_{p2}}=&-(1+\Delta )\frac{{{l}_{p2}}}{{{\varepsilon }^{2}}}{{\mu }_{p2}}-(1+\Delta )\frac{{{l}_{p1}}{{l}_{p2}}}{{{\varepsilon }^{2}}}{{\mu }_{p1}} \\
 &+(1+\Delta ){{l}_{p2}}{{\dot{x}}_{p2}}+\varepsilon {{\gamma }_{p1}}\label{kesai2 dao}
\end{aligned}
\end{equation}
where
\begin{align*}
\Delta =\text{sat}'\frac{m_{0}^{2}m}{{{m}_{0}}-m},\,\,
{{\gamma }_{p1}}=\frac{{{{\dot{d}}}_{f}}}{m}+\text{sat}'\frac{m_{0}^{2}m}{{{m}_{0}}-m}({{k}_{1}}{{x}_{p2}}+{{k}_{2}}{{\dot{\hat{x}}}_{p2}})
\end{align*}
Since $\left| \text{sat}' \right|\le 1$, $1+\Delta >0$, $\forall t\in \left[ {{T}_{1}},{{t}_{2}} \right]$. 

In \eqref{mu1} and \eqref{kesai2 dao}, there exists ${{\varepsilon }_{2}}\in \left( 0,{{\varepsilon }_{1}} \right]$ such that any $\varepsilon \in \left( 0,{{\varepsilon }_{2}} \right]$ and ${{T}_{2}}\in \left( {{T}_{1}},{{t}_{1}} \right]$,
\begin{equation}
{{\mu }_{p2}}=O(\varepsilon ), \,\forall t\in \left[ {{T}_{2}},{{t}_{2}} \right] \label{mu2}
\end{equation}

\begin{figure*}[t!]
\centering
 {\includegraphics[width=17.5cm]{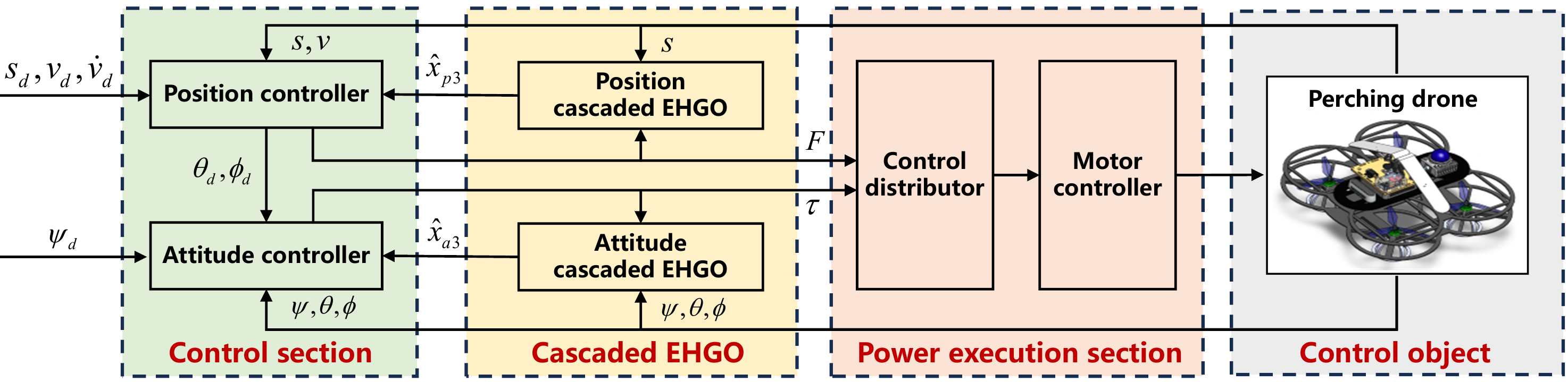}}
\caption{Control system architecture and data flow in perching drone.}
\label{控制框图}
\end{figure*}

Based on the definition of ${{\xi }_{p2}}$ and ${{{\hat{x}}}_{p3}}$ is restricted by a positive number independent of $\varepsilon $ within the time interval $\left[ {{T}_{1}},{{t}_{2}} \right]$. Since an appropriate saturation limit $B$ can be selected, it ensures that the control ${{m}_{0}}\left( K{{{\hat{x}}}_{p}}-{{{\hat{x}}}_{p3}} \right)$ is outside the saturation region within the time interval $\left[ {{T}_{1}},{{t}_{2}} \right]$. Therefore, $\forall t\in \left[ {{T}_{2}},{{t}_{2}} \right]$,
\begin{equation}\label{xp2dao}
{{\dot{x}}_{p2}}=\frac{1}{\varepsilon }{{\mu }_{p2}}+{{\gamma }_{p2}},
\end{equation}
where ${{\gamma }_{p2}}=K\hat{x}_{p}$. In \eqref{mu1} and \eqref{mu2}, both ${{\gamma }_{p2}}$ and ${{\dot{\gamma }}_{p2}}$ are restricted by a positive number independent of $\varepsilon $ within the time interval $\left[ {{T}_{2}},{{t}_{2}} \right]$.

To establish that ${{\gamma }_{p2}}$ satisfies $O({{\varepsilon }^{2}})$, \eqref{miu1 dao}, \eqref{kesai2 dao} and \eqref{xp2dao} are combined as follows
\begin{equation}
{{\dot{\mu }}_{p}}=\frac{1}{\varepsilon }{{\alpha }_{1}}{{\mu }_{p}}+{{\alpha }_{2}}{{\gamma }_{p2}}+\varepsilon {{\alpha }_{3}}{{\gamma }_{p1}} \label{mu dao}
\end{equation}
where
\[{{\alpha }_{1}}=\left[ \begin{matrix}
 -{{l}_{1}} & 1 \\
 -{{l}_{1}}{{l}_{2}}(1+\Delta ) & 0 \\
\end{matrix} \right],{{\alpha }_{2}}=\left[ \begin{matrix}
 1 \\
 {{l}_{2}}(1+\Delta ) \\
\end{matrix} \right],{{\alpha }_{3}}=\left[ \begin{matrix}
 0 \\
 1 \\
\end{matrix} \right]\]
The matrix ${\alpha }_{1}$ is Hurwitz and invertible, and ${{\alpha }_{1}}^{-1}{{\alpha }_{2}}={{\left[ \begin{matrix}
 -l_{1}^{-1} & 0 \\
\end{matrix} \right]}^\text{T}}$, subtracting the state transformation $\Phi ={{\mu }_{p}} +\varepsilon \alpha _{1}^{-1}{{\alpha }_{2}}{{\gamma }_{p2}}$ into \eqref{mu dao} :
\begin{equation}
\dot{\Phi }=\frac{1}{\varepsilon }{{\alpha }_{1}}\Phi +\varepsilon (\alpha _{1}^{-1}{{\alpha }_{2}}{{\dot{\gamma }}_{p2}}+{{\alpha }_{3}}{{\gamma }_{p1}}) \label{kesai dao}
\end{equation}
With the boundedness of ${{\gamma }_{p1}}$ and ${{\dot{\gamma }}_{p2}}$, there exists ${{\varepsilon }_{3}}\in \left( 0,{{\varepsilon }_{2}} \right]$ in  \eqref{kesai dao} such that any $\varepsilon \in \left( 0,{{\varepsilon }_{3}} \right]$ and ${{T}_{3}}\in \left( {{T}_{2}},{{t}_{1}} \right]$,
\begin{equation}
\left\| \Phi \right\|=O({{\varepsilon }^{2}}),\forall t\in \left[ {{T}_{3}},{{t}_{2}} \right]
\end{equation}
The dynamics of ${{\mu }_{p2}}$ can be expressed as follows:
\begin{equation}
{{\mu }_{p2}}=\alpha _{3}^{T}{{\mu }_{p}}=\alpha _{3}^{T }\Phi -\varepsilon \alpha _{3}^{T}\alpha _{1}^{-1}{{\alpha }_{2}}{{\gamma }_{p2}}
\end{equation}
Since $\alpha _{3}^{\text{T} }\alpha _{1}^{-1}{{\alpha }_{2}}=\left[ \begin{matrix}
 0 & 1 \\
\end{matrix} \right]{{\left[ \begin{matrix}
 -l_{1}^{-1} & 0 \\
\end{matrix} \right]}^{\text{T} }}=0$, ${{\mu }_{P2}}=\alpha _{3}^{\text{T} }\Phi $, there exists that ${{\varepsilon }_{3}}\in \left( 0,{{\varepsilon }_{2}} \right]$ such that any $\varepsilon \in \left( 0,{{\varepsilon }_{3}} \right)$ and ${{T}_{3}}\in \left( {{T}_{2}},{{t}_{1}} \right)$,
\begin{equation}
\left| {{x}_{3}}-{{{\hat{x}}}_{3}} \right|=O\left( \varepsilon \right),\forall t\in \left[ {{T}_{3}},{{t}_{2}} \right]
\end{equation}

\textit{Convergence of ${{x}_{P}}$}: Based on the definition of $V\left( {{x}_{P}} \right)$, \eqref{sigma budengshi} can be expressed as follows: 
\begin{equation}
\begin{cases}
{{\sigma }_{1}}{{\left\| {{x}_{P}} \right\|}^{2}}\le V\left( {{x}_{P}} \right)\le {{\sigma }_{2}}{{\left\| {{x}_{P}} 
\right\|}^{2}}\\\left| \frac{\partial V}{\partial {{x}_{P}}}({{x}_{P}}) \right|\le 2{{\sigma }_{2}}\left\| {{x}_{P}} \right\| \label{sigma budengshi}\\
\end{cases}
\end{equation}
where ${{\sigma }_{1}}$ and ${{\sigma }_{2}}$ are the smallest and largest eigenvalues of the symmetric matrix $P$, respectively. The dynamics of $V\left( {{x}_{P}} \right)$ within the time interval $\left[ {{T}_{3}},{{t}_{2}} \right]$ can be given as follows:
\begin{equation}\label{vx fangsuo}
\begin{aligned}
\dot{V}({{x}_{P}}) &= {{x}_{P2}} \frac{\partial V}{\partial {{x}_{P1}}}({{x}_{P}}) + \left( K\hat{x}_{P} + {{x}_{P3}} - \hat{x}_{P3} \right) \frac{\partial V}{\partial {{x}_{P2}}}({{x}_{P}}) \\
&\le -\| {{x}_{P}} \|^2 + 2\sigma_2 |{{\gamma}_{P3}}| \|{{x}_{P}} \| 
\end{aligned}
\end{equation}
where
\begin{equation}
{{\gamma }_{P3}}={{x}_{P3}}-{{\hat{x}}_{P3}}+K{\hat{x}}_{P}-K{{x}_{p}}
\end{equation}
For the stability of the observer, any $\varepsilon \in (0,{{\varepsilon }_{3}})$ and $t\in \left[ {{T}_{3}},{{t}_{2}} \right]$, $\left| {{\gamma }_{p3}} \right|=O(\varepsilon )$. In \eqref{lamuda assum} and 
\eqref{sigma budengshi}, $\sqrt{\frac{{{\lambda }_{0}}}{{{\sigma }_{2}}}}\le \left\| {x}_{p} \right\|\le \sqrt{\frac{{{\lambda }_{0}}+1}{{{\sigma }_{1}}},}t\in \left[ 0,{{t}_{2}} \right]$. Therefore, there exists ${{\varepsilon }_{4}}\in \left( 0,{{\varepsilon }_{3}} \right]$ such that any $\varepsilon \in \left( 0,{{\varepsilon }_{4}} \right)$, 
\begin{equation}
\dot{V}({x}_{p})\le 0,t\in \left[ {{t}_{1}},{{t}_{2}} \right]
\end{equation}
This contradicts \eqref{lamuda assum}, thereby the lemma is proved.

Based on the lemma, we are in a position to state the proof of the theorem.

\textit{Proof of the theorem}: By the lemma, for any $\varepsilon \in \left( 0,\varepsilon\right)$, ${x}_{p}(t)\in {{\Omega }_{1}},\forall t\in \left[ 0,\infty \right)$. By \eqref{miu1 dao}, within the time interval $t\in \left[ 0,\infty \right)$, the term ${{{\dot{x}}}_{p2}}$ is bounded. Consequently there exists $\varepsilon _{1}^{*}\in \left( 0,\varepsilon \right)$ and ${{T}^{*}}\in \left( 0,T \right)$ such that for any $\varepsilon \in \left( 0,\varepsilon _{1}^{*} \right)$ and $t\in \left[ {{T}^{*}},\infty \right)$, $\left| {{x}_{p2}}-{{{\hat{x}}}_{p2}} \right|=O\left( \varepsilon \right)$. By \eqref{mu dao}, for any $T>0$, there exists $\varepsilon _{2}^{*}\in \left( 0,\varepsilon _{1}^{*} \right]$ such that for any $\varepsilon \in \left( 0,\varepsilon _{2}^{*} \right)$, $\left| {{x}_{p3}}-{{{\hat{x}}}_{p3}} \right|=O\left( \varepsilon \right)$ holds. By \eqref{vx fangsuo}, there exists ${{\varepsilon }^{*}}\in \left( 0,\varepsilon _{2}^{*} \right]$ such that \eqref{x shoulian} holds. Therefore, the theorem is proved.

This new cascaded observer only requires the control gain symbols to be consistent, without the need for precise estimation of the gain magnitude, thereby reducing the dependence on the system model. The system adopts a multi-stage first-order observer cascaded structure and maintains stability through internal dynamic regulation, greatly simplifying the parameter tuning process.The block diagram illustrating the drone control system architecture is presented in \cref{控制框图}. In this controller, a cascaded EHGO is integrated to continuously monitor and compensate for disturbances, thereby enhancing the stability and robustness of the controller.



\begin{figure*}[t!]
\centering
 {\includegraphics[width=17.5cm]{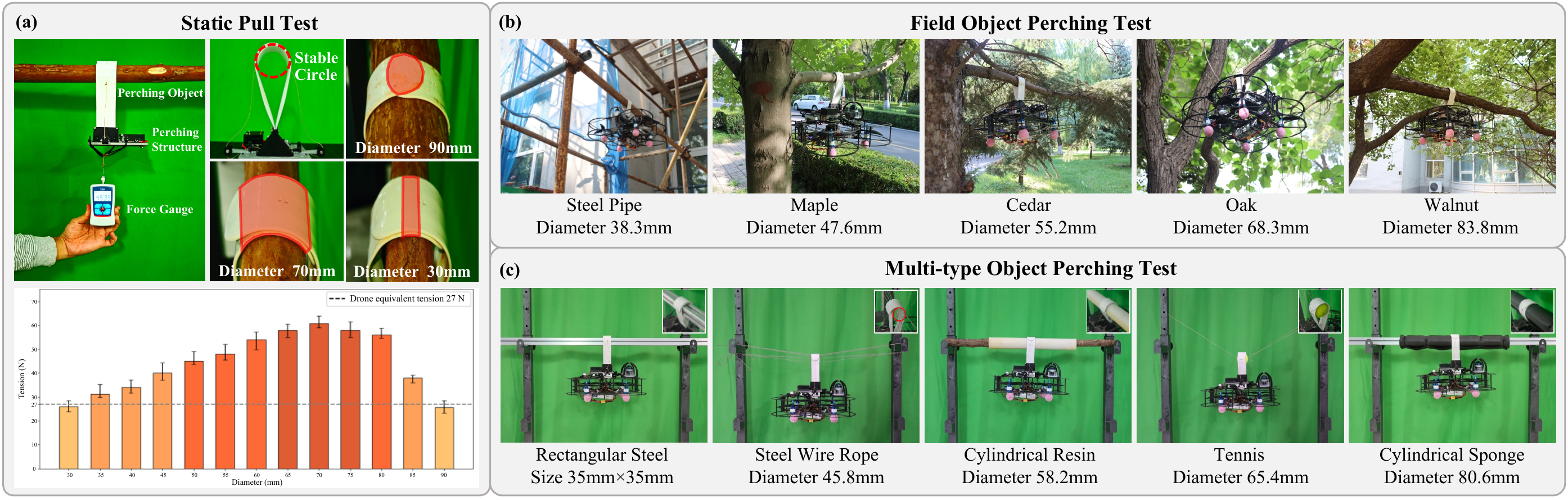}}
\caption{(a) Static Pull test: Use a tensiometer to test and fit the corresponding relationship between the diameter of the perch and the tensile force that the mechanism can withstand. (b) Field object perching test: Test the perching ability of the mechanism on the construction steel pipe and four different types and thicknesses of tree branches. (c) Multi-type object perching test: Test the perching ability of the mechanism on five different materials and sizes of objects. (Note: The diameter of the steel wire rope refers to the geometric diameter formed between the ropes in a stable perching state.)}
\label{栖息测试}
\end{figure*}

\section{Experiment Results}


In this section, real-world institutional tests and flight experiments are conducted to demonstrate the mechanical advantages of perching drones and the controller. In the experiments, the drone acquires and records the real-time position through the Mid-360 laser radar and the attitude through the IMU from the Pixhaw 6C. To demonstrate the effectiveness and superiority of the developed EHGO-based method, the well-tuned  classical PID and standard EHGO controllers are implemented for comparison.

\subsection{Perching suitability testing and analysis}

The perching performance evaluation of the mechanism is presented in \cref{栖息测试}, with the static pull test detailed in Figure (a). Thirteen wooden specimens exhibiting similar surface roughness are selected at uniform diameter intervals ranging from 30 mm to 90 mm as experimental substrates. Each specimen undergoes five repeated trials. Following complete attachment of the perching mechanism to the substrate, a GYJ-F100SS digital force gauge equipped with a peak-hold function is employed to apply a constant pulling speed. The maximum tensile force is recorded at the moment of mechanism detachment. Data analysis reveals that the relationship between the tensile force the mechanism can withstand and the diameter of the perching object follows a monomodal asymmetric distribution. As the diameter of the perching object approaches the designed stable contact diameter, the contact area between the perching belts increases, resulting in enhanced tensile resistance. Based on the results of the test data and the drone's weight of 2.7 kg, the operational diameter range of the mechanism is defined as 35 mm to 85 mm. Utilizing this defined diameter range, field object perching test and multi-type object perching test are conducted to evaluate the mechanism's adaptability. Experimental results demonstrate that the perching mechanism exhibits reliable performance across a variety of indoor and outdoor perching substrates within the specified diameter range, as shown in Figures (b) and (c).
\subsection{Perching, Escaping and Landing}
\begin{figure*}[t!]
\centering
 {\includegraphics[width=17.2cm]{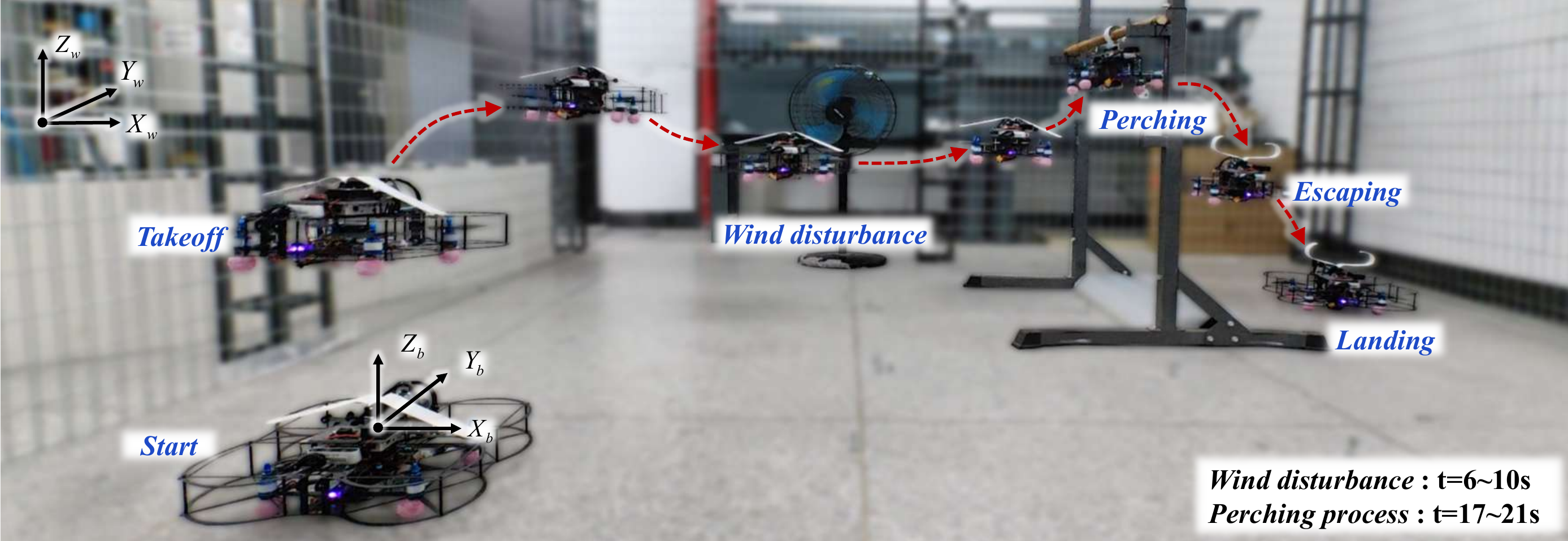}}
\caption{Perching drone flight sequence. (Note: Illustration of the drone’s operational sequence includes takeoff, perching, escaping, and landing. During the process, an electric fan generates the wind disturbance to evaluate the effectiveness of the cascaded EHGO in maintaining stability.)}
\label{实验轨迹}
\end{figure*}

\begin{figure}[t!] 
	\centerline{\includegraphics[width=7.5cm]{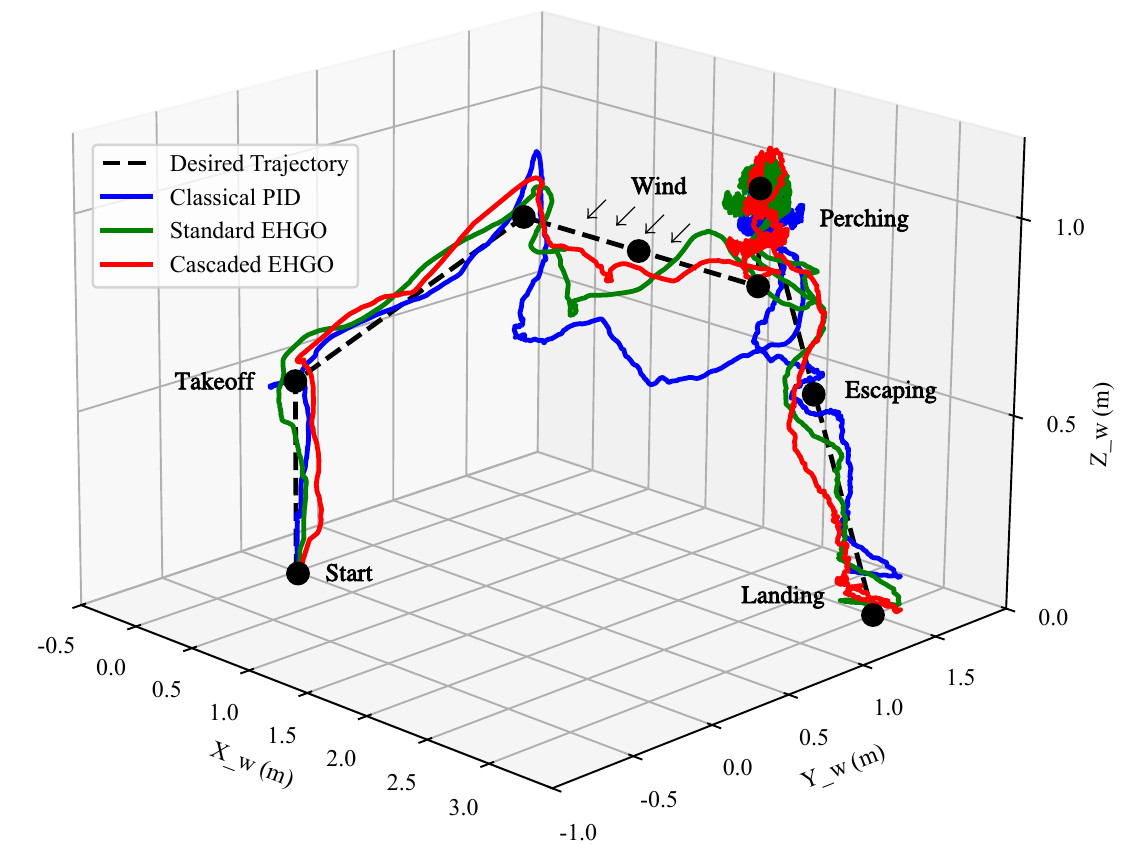}}
 \caption{3D trajectory diagram of the drone.}
\label{轨迹位置}
\end{figure}

The perching mission process is depicted in \cref{实验轨迹}, with the corresponding trajectory representation in \cref{轨迹位置}. The drone takes off and follows the pre-determined flight path. When the upward camera detects the perching target, it triggers the perching mechanism. After a stable perching, the drone shuts down its propellers. Upon receiving the escape command, the drone releases the perching mechanism, escapes and lands.

\begin{figure}[t!] 
	\centerline{\includegraphics[width=8.2cm]{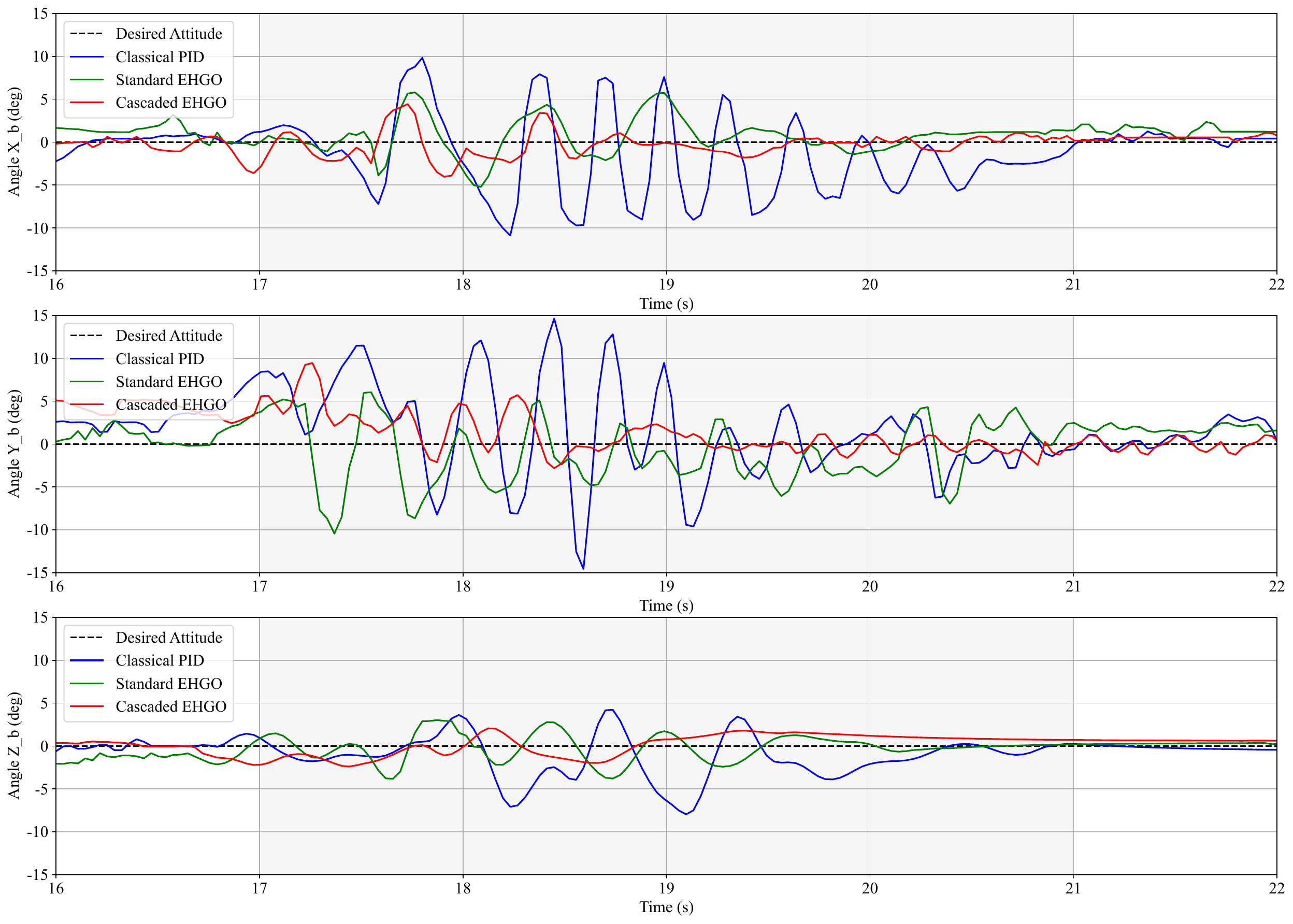}}
 \caption{Comparison of the attitude during the perching process. (Note: t=17.5s, the perching mechanism collides with the target.)}
\label{角度}
\end{figure}\normalsize

\begin{figure}[ht] 
	\centerline{\includegraphics[width=8.2cm]{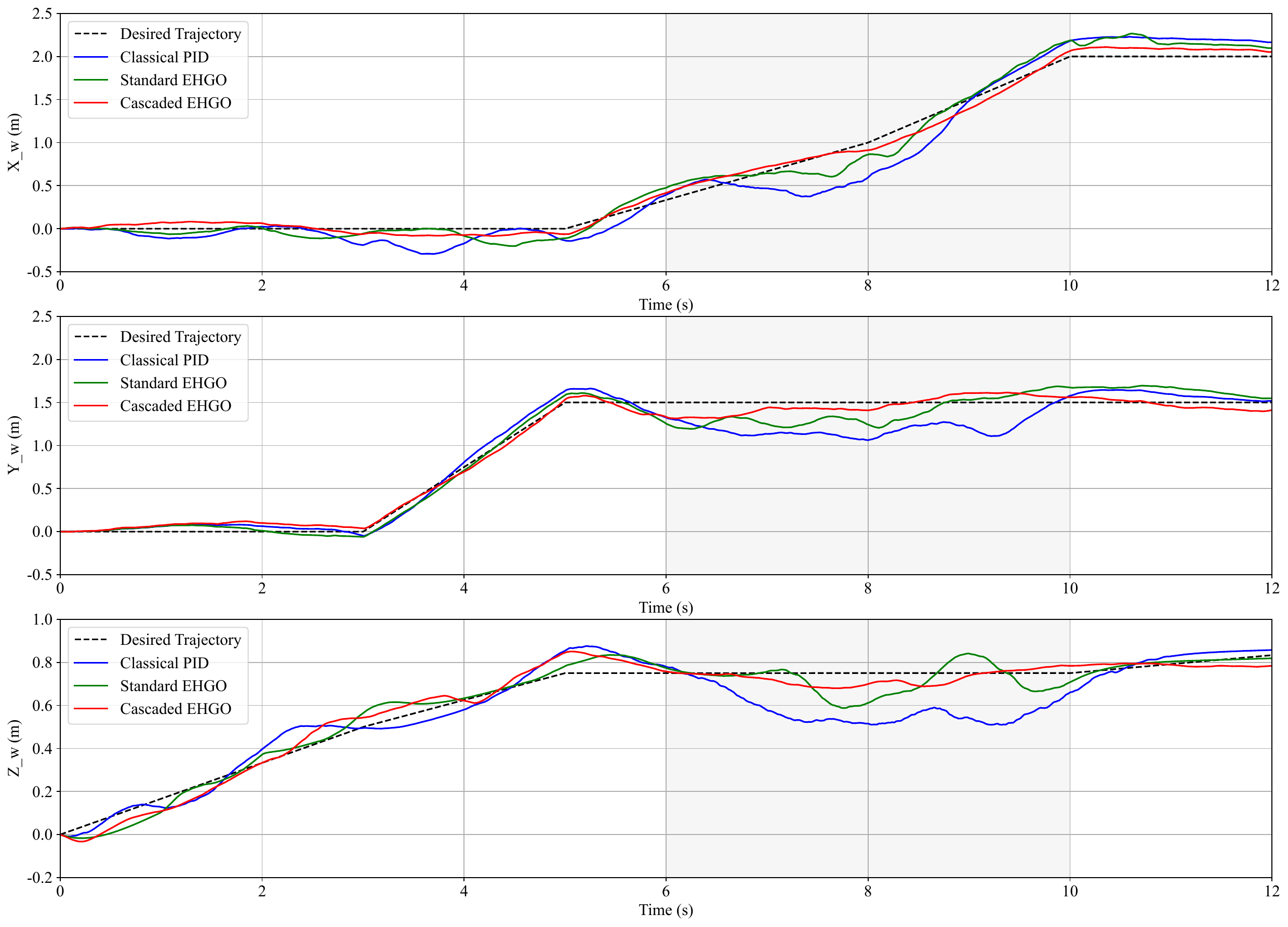}}
 \caption{Comparison of the wind disturbance impacts on drone stability. (Note: t=6s-10s, the greatest wind disturbance.)}
\label{位置}
\end{figure}\normalsize



\normalsize

\subsection{Perhing Disturbance Experiment}
When the drone impacts a perching target, it induces shifts in the center of mass and moment of inertia, resulting in discrepancies between the actual and predicted models. The experiment evaluates the performance of classical PID, standard EHGO, and cascaded EHGO in suppressing attitude disturbances in the drone (\cref{角度}), demonstrating that the cascaded EHGO approach achieves the most effective suppression of jitter.
The ${Y}_{b}$ axis undergoes a higher level of impact disturbance during the perching process, thereby providing a more rigorous scenario for evaluating the performance of the cascaded EHGO. In Table \ref{errors}, compared to the standard EHGO and classical PID, the cascaded EHGO achieves a 25.1\% and 67.6\% reduction in average error, respectively, along with a 79.3\% and 83.6\% decrease in standard error.

\subsection{Wind Disturbance Experiment}

To simulate the impact of the strong wind disturbance on the drone during outdoor flight, a fan with a maximum wind speed of 10 m/s is used to generate a localized wind field. In \cref{位置}, the drone equipped with the cascaded EHGO exhibits notably reduces positional errors during wind disturbances and achieves faster trajectory recovery compared to the standard EHGO and classical PID. The ${Y}_{w}$ axis is more significantly affected by the wind disturbance during flight, thereby providing a more effective scenario for evaluating the performance of the cascaded EHGO. As presented in Table 1, compared to the standard EHGO and classical PID, the cascaded EHGO algorithm achieves a 17.7\% and 21.4\% reduction in average error, respectively, along with a 22.31\% and 33.3\% decrease in standard error.


\begin{table}[t!]
\centering
\renewcommand{\arraystretch}{1} 
\setlength{\tabcolsep}{6pt}
\caption{QUANTITATIVE COMPARISON OF ERRORS}
\label{errors}
\begin{tabular}{llcc}
\hline
\textbf{Attitude loop stability test}& & \textbf{Mean} & \textbf{STD} \\
\hline
\multirow{2}{*}{${X}_{b}$-axis angular error (deg)} 
 & Classical PID & 4.84 & 1.60 \\
 & Standard EHGO & 3.31 & 1.09 \\
 & Cascaded EHGO & 2.36 & 0.24 \\
\hline
\multirow{2}{*}{${Y}_{b}$-axis angular error (deg)} 
 & Classical PID & 4.14 & 1.10 \\
 & Standard EHGO & 1.79 & 0.87 \\
 & Cascaded EHGO & 1.34 & 0.18 \\
\hline
\multirow{2}{*}{${Z}_{b}$-axis angular error (deg)} 
 & Classical PID & 2.19 & 0.89 \\
 & Standard EHGO & 1.37 & 0.25 \\
 & Cascaded EHGO & 1.11 & 0.18 \\
\hline
\textbf{Position loop stability test}& & \textbf{Mean} & \textbf{STD} \\
\hline
\multirow{2}{*}{${X}_{w}$-axis position error (cm)} 
 & Classical PID & 7.44 & 10.22 \\
 & Standard EHGO & 7.28 & 8.74 \\
 & Cascaded EHGO & 5.67 & 6.38 \\
\hline
\multirow{2}{*}{${Y}_{w}$-axis position error (cm)} 
 & Classical PID & 9.16 & 12.22 \\
 & Standard EHGO & 8.75 & 10.49 \\
 & Cascaded EHGO & 7.20 & 8.15 \\
\hline
\multirow{2}{*}{${Z}_{w}$-axis position error (cm)} 
 & Classical PID & 8.35 & 9.64 \\
 & Standard EHGO & 7.61 & 8.31 \\
 & Cascaded EHGO & 3.33 & 3.99 \\
\hline
\end{tabular}
\end{table}

\section{Conclusion and Discussion}

Inspired by the Venus flytrap, this paper presents a compliant perching mechanism that integrates proactiveness, rapid response, and adaptability. This mechanism can achieve a stable closure within 100 ms. To address disturbances encountered during flight and perching, a new cascaded EHGO is developed. This observer effectively monitors and compensates for instantaneous impacts and external wind disturbances. Experimental results validate the effectiveness of the proposed perching drone, demonstrating its rapid perching capability and stable motion performance even under challenging conditions. In the future, the rapidity of this perching structure will be utilized to achieve dynamic target grasping and focus on the energy consumption reduction brought by the perching mechanism, further enhancing the flexibility and functional diversity of drones.

\bibliographystyle{IEEEtran}
\bibliography{ref}

\end{document}